\documentclass[10pt,twocolumn,letterpaper]{article}

\usepackage[pagenumbers]{cvpr}              
\usepackage{multirow}
\usepackage{colortbl}
\definecolor{up}{RGB}{135,150,235}
\usepackage{tikz} 
\usepackage{subcaption}
\usepackage{graphicx}
\usepackage[dvipsnames]{xcolor} 
\usepackage{overpic}    
\usepackage{amssymb}

\usetikzlibrary{calc}

\definecolor{cvprblue}{rgb}{0.21,0.49,0.74}
\usepackage[breaklinks,colorlinks,allcolors=cvprblue]{hyperref}

\def\confName{CVPR}
\def\confYear{2026}

\title{RoSAMDepth: Robust Self-supervised Depth Estimation Leveraging Segment Anything Model}

\author{
Xuanang Gao$^{1,3}$ \quad
Zhiwei Ning$^{1,3}$ \quad
 Gengming Zhang$^{1,3}$ \quad
Jiaxi Cao$^{1,3}$ \quad
Runze Yang$^{1,4}$ \quad \\
Zhonglong Zheng$^{5}$ \quad
Jie Yang$^{1,2,3}$ \quad
Rong Xiao$^{6}$ \quad
Wei Liu$^{1,2,3}$ \thanks{Corresponding Author: Wei Liu}
\\
\footnotesize{$^1$School of Automation and Intelligent Sensing, Shanghai Jiao Tong University} 
\footnotesize{$^2$Institute of Medical Robotics, Shanghai Jiao Tong University} \\
\footnotesize{$^3$Institute of Image Processing and Pattern Recognition, Shanghai Jiao Tong University} 
\footnotesize{$^4$School of Computing, Macquarie University} \\
\footnotesize{$^5$School of Computer Science, Zhejiang Normal University} 
\footnotesize{$^6$IntelliFusion Inc.} \\
\vspace{-4mm}
}

\hypersetup{
  pdftitle={RoSAMDepth: Robust Self-supervised Depth Estimation Leveraging Segment Anything Model},
  pdfauthor={Xuanang Gao; Zhiwei Ning; Gengming Zhang; Jiaxi Cao; Runze Yang; Zhonglong Zheng; Jie Yang; Rong Xiao; Wei Liu}
}

\begin{document}
\maketitle
\begin{abstract}
Robust depth estimation aims to maintain high-quality depths across diverse conditions. 
However, most existing methods estimate depth without taking into account the object-level information. 
As a result, the predicted depth may easily deviate within objects and become blurred under adverse conditions. 
To overcome this weakness, we propose RoSAMDepth, a novel framework that can assist robust self-supervised depth estimation in leveraging rich and diverse object-level priors from the Segment Anything Model (SAM). 
We focus on incorporating object-level information across three key aspects: a segment-guided representation contrasting method that injects object-level awareness into the feature representation space; an adaptive regional outlier masking strategy combined with a regional Gaussian likelihood loss that enforces regional depth smoothness; and an object-level reliability estimation strategy that mitigates the influence of unreliable supervision. 
Extensive experiments across multiple datasets and diverse weather conditions demonstrate that our method produces sharper, more accurate depth predictions, consistently outperforming state-of-the-art methods. Our code is available at \href{https://github.com/xagao/RoSAMDepth}{https://github.com/xagao/RoSAMDepth}.

\end{abstract}
    
\vspace{-0.3em}
\section{Introduction}
\label{sec:intro}

\begin{figure}
    \centering
    \includegraphics[width=0.96\linewidth]{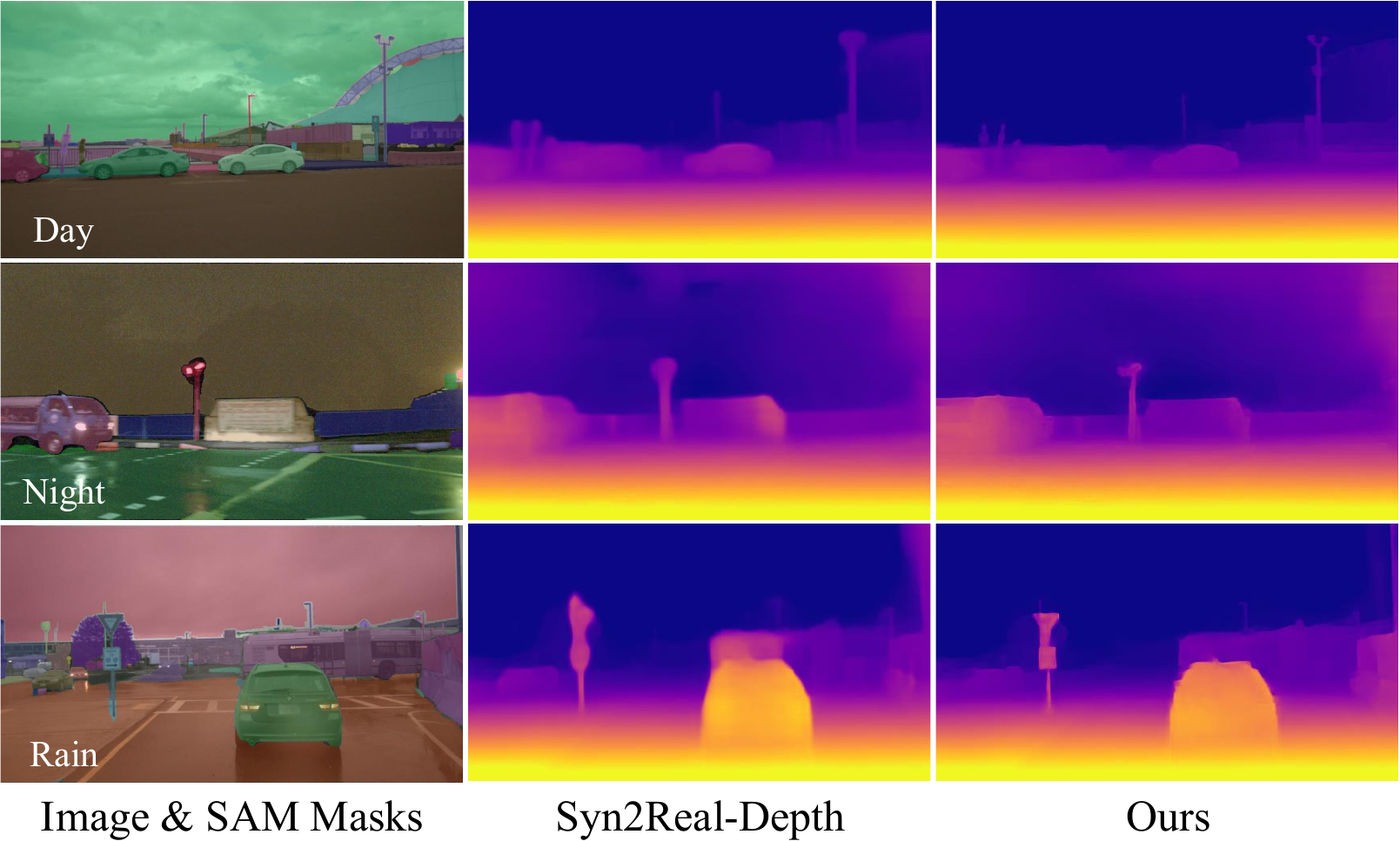}
    \caption{Our RoSAMDepth utilizes object-level information from SAM \cite{kirillov2023segment} to generate sharp and accurate depths across diverse conditions.}
    \vspace{-1.5em}
    \label{fig:fig1}
\end{figure}

Monocular depth estimation is a fundamental task in computer vision, crucial for downstream applications, such as autonomous driving, robotics, and AR/VR. 
Supervised methods \cite{bhat2021adabins, fu2018deep, lee2019big, li2023learning, ranftl2021vision} achieve high accuracy but rely on expensive ground-truth (GT) depths for training. 
To alleviate this dependency, self-supervised methods leverage geometrical constraints from stereo pairs or monocular videos to learn depth estimation without explicit annotations \cite{zhou2017unsupervised, godard2019digging, zhang2023lite, he2022ra, bian2021unsupervised, gasperini2023robust, moon2024ground, watson2021temporal}. 

Despite their success under standard daytime conditions, self-supervised methods often degrade in challenging conditions such as nighttime or adverse weather \cite{gasperini2023robust, saunders2023self, mao2024stealing, yan2025synthetic}, where the photometric consistency breaks. 
Early works address a specific condition (e.g., nighttime), using adversarial learning or feature disentanglement for domain representations \cite{spencer2020defeat,liu2021self,vankadari2020unsupervised}. 
Recent methods \cite{gasperini2023robust,saunders2023self,mao2024stealing,yan2025synthetic} focus on unified robustness across diverse degradations by synthesizing paired data with teacher–student learning or further adapting synthetic-trained teachers to real degraded data.

However, existing methods typically treat degraded regions uniformly, without considering the semantic or object-level differences that significantly influence visual perception and depth estimation. 
As shown in \cref{fig:fig1}, ignoring object-level priors can lead to blurred depth boundaries and unstable predictions under complex environments or diverse conditions. 
To better estimate depth across diverse real data, it is important to understand the object spatial relationships and assess the reliability of depth cues. 
Meanwhile, the predicted depth should also be smooth only within the same continuous object region, while allowing sharp depth boundaries near object edges. 
Thus, incorporating object-level information is key to refining robust self-supervised depth estimation under adverse conditions. 

Indeed, prior methods exploit semantic segmentation to incorporate object information for depth estimation under standard conditions \cite{jung2021fine, ma2022towards}. 
However, semantic segmentation is inherently limited, as it cannot distinguish instances within the same class, which may have drastically different depths. It is also constrained by a limited set of predefined classes, often failing to recognize novel objects in the open world. 
Moreover, segmentation models trained on specific domains or weather conditions generalize poorly across diverse scenarios. 
In comparison, the Segment Anything Model (SAM) \cite{kirillov2023segment} offers a better option. SAM is a general-purpose image segmentation model pre-trained on massive and diverse datasets. It separates individual instances and demonstrates strong zero-shot robustness to unseen objects and various weather conditions \cite{kirillov2023segment, qiao2023robustness, shan2023robustness}. 

Building upon these advantages, we propose to integrate SAM for additional object-level information to enhance robust self-supervised depth estimation. 
Our framework jointly improves feature representation and depth learning. 
Specifically, for representation learning, we propose segment-guided representation contrasting (SRC) that exploits SAM masks for implicitly embedding object-level structural awareness into the feature space. 
Furthermore, to improve supervision for depth learning, we identify key limitations in the conventional smoothness loss and the pixel-level pseudo-label quality indicator. In response, we introduce two strategies: an adaptive regional outlier masking (AROM) strategy paired with a regional Gaussian likelihood loss to rectify inconsistent depth outliers; and an object-level reliability estimation (ORE) strategy that mitigates the influence of unreliable supervision signals. 
Our method significantly outperforms previous methods in diverse conditions through different real-world datasets \cite{caesar2020nuscenes, maddern20171}. In summary, our contributions are as follows:
\begin{itemize}
    \item To the best of our knowledge, we are the first to effectively integrate SAM into robust self-supervised depth estimations, using only pre-generated SAM masks during the training as object-level priors. 
    \item We propose three novel designs to take full advantage of SAM's object-level priors: segment-guided representation contrasting (SRC), adaptive regional outlier masking (AROM) strategy combined with a regional Gaussian constraint, and object-level reliability estimation (ORE) strategy. These components provide efficient guidance to train the model and lead to improved depth. 
    \item Extensive experiments on the nuScenes \cite{caesar2020nuscenes} and Oxford Robotcar \cite{maddern20171} benchmarks demonstrate that our method consistently outperforms state-of-the-art approaches, delivering sharper and more accurate depth under diverse real-world conditions.
\end{itemize}

\vspace{-0.3em}
\section{Related Work}
\label{sec:related work}
\subsection{Self-supervised Monocular Depth Estimation}
Self-supervised monocular depth estimation methods use multi-view images for training and learn to estimate the depth from a single input image with photometric reconstruction consistency as supervision. SfM-Learner \cite{zhou2017unsupervised} firstly introduces an unsupervised paradigm that relies on photometric consistency, formulating depth estimation as a view synthesis task. Monodepth2 \cite{godard2019digging} introduces the per-pixel photometric loss with an auto-mask and a full-resolution sampling for training the model to reduce the impact of moving objects. Several methods \cite{guizilini2020semantically,feng2022disentangling,jung2021fine, ma2022towards} use extra semantic information to improve performance, and the frameworks explored in \cite{yin2018geonet, li2021unsupervised} jointly learn depth, camera pose, and optical flow. Additionally, the multi-frame monocular depth estimation is handled in \cite{guizilini2022multi, watson2021temporal}, which predicts more accurate depths by taking two frames of a monocular video as the input. 
\subsection{Depth Estimation under Adverse Conditions} 
Robust depth estimation under adverse conditions, such as rain, night, and fog, remains challenging and actively studied. 
Early approaches focus on nighttime. DeFeat-Net \cite{spencer2020defeat} jointly learns depth and cross-domain features, ADIDS \cite{liu2021self} utilizes a generative adversarial network (GAN) to generate paired nighttime images and splits features into private and invariant sub-spaces, and ADFA \cite{vankadari2020unsupervised} employs a GAN-based discriminator with adversarial training to align nighttime and daytime features. 
Later methods like RNW \cite{wang2021regularizing} and STEPS \cite{zheng2023steps} integrate image enhancement to improve depth quality, and WSGD \cite{vankadari2023sun} combines denoising with a lighting change decoder to predict per-pixel changes. 

More recent methods extend beyond nighttime to tackle multiple conditions by exploiting generative techniques such as GANs and diffusion models \cite{zheng2020forkgan, zhu2017unpaired, rombach2022high, zhang2023adding}. Md4all \cite{gasperini2023robust}, Robust-Depth \cite{saunders2023self}, and WeatherDepth \cite{wang2024weatherdepth} enhance robustness through GAN-based weather augmentations integrated into teacher-student frameworks. Complementary to these approaches, DM-MDE \cite{tosi2024diffusion} incorporates stable diffusion priors, and Syn2RealDepth \cite{yan2025synthetic} proposes an auxiliary space and a real adaptation stage.  
Our key insight is that the fidelity of synthetic-to-real adaptation can be significantly improved with explicit object-level guidance. We operationalize this by incorporating cues from SAM, which enhance depth estimation in two ways: 1) enabling the model learning object-aware perception, and 2) providing object-level priors for supervision signals refinement. 

\begin{figure*}[!thbp]
    \centering
    \includegraphics[width=0.97\linewidth]{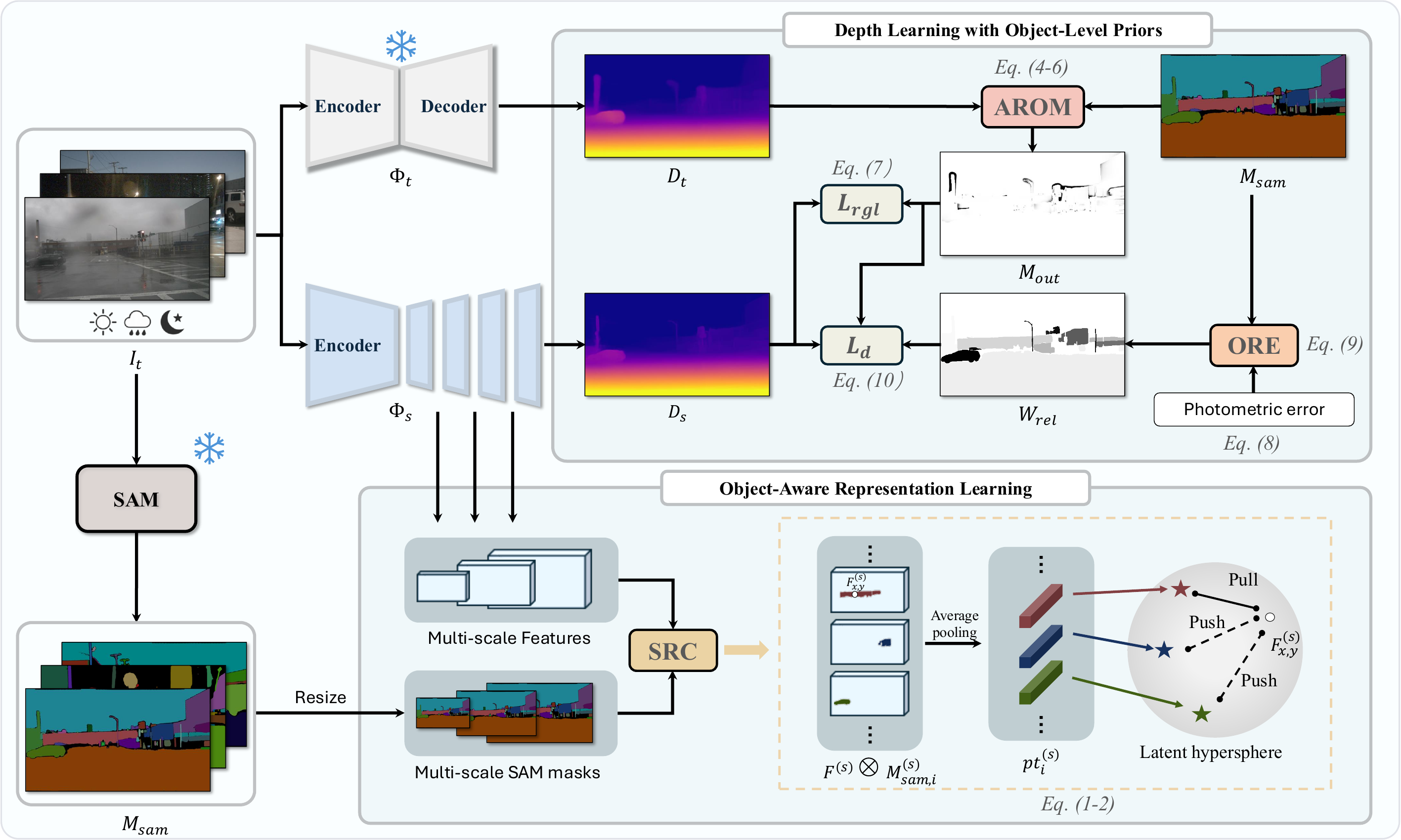}
    \caption{Illustration of the proposed method. Given the SAM masks, we leverage object-level priors to improve depth estimation $D_s$ in two complementary ways: 1) Object-aware representation learning: SRC injects segment priors into multi-scale features. 2) Depth learning with object-level priors, AROM generates an outlier-suppressed mask $M_{out}$, which is paired with $L_{rgl}$ to enforce consistent depth within each region. Concurrently, ORE estimates object-level reliability to form a reliability-weighted map $W_{rel}$, combined with $M_{out}$ to modulate the distillation loss $L_d$.}
    \vspace{-1.0em}
    \label{fig:framework}
\end{figure*}

\vspace{-0.2em}
\section{Method}
\label{sec:method}
We target accurate and realistic depth estimation across various conditions, including both standard (e.g., daytime) and adverse conditions (e.g., rain, nighttime, etc.). 
Our pipeline, illustrated in \cref{fig:framework}, is built around two main objectives: 1) learning object-aware representation, and 2) improving depth learning signals with object-level priors. 
To achieve the first objective, we propose segment-guided representation contrasting (SRC), which enables the depth network to implicitly learn the concept of segmentation at the feature level (\cref{subsectin:feature_level}). 
To realize the second objective, we introduce two key strategies: an adaptive regional outlier masking (AROM) strategy, which is combined with a regional Gaussian likelihood loss to enforce object-aware depth smoothness, and an object-level reliability estimation (ORE) strategy to focus self-supervision on reliable regions (\cref{subsection:depth level}). 
The training of the teacher depth network ($\Phi_t$) is detailed in \cref{subsection:preliminaries}. 

\subsection{Preliminaries}
\label{subsection:preliminaries}
\noindent \textbf{Self-supervised Depth Estimation} 
Given a target image $I_t\in \mathbb{R}^{H\times W}$ and an adjacent source frame $I_s \in \mathbb{R}^{H\times W}$, self-supervised monocular depth estimation learns to predict a dense depth map $D \in \mathbb{R}^{H\times W}$ for $I_t$. 
Instead of ground-truth labels, supervision is provided by a photometric reprojection loss that measures the consistency between $I_t$ and its reconstruction $I_{t'}$, obtained by warping $I_s$ using the predicted $D$ and the estimated relative pose. 

\noindent \textbf{Robust Teacher Depth}
We build upon the synthetic-to-real training paradigm introduced in~\cite{yan2025synthetic}, which consists of a synthetic adaptation stage followed by a real adaptation stage. 
Our improvements focus on the real adaptation stage, where the model encounters the challenges of real-world data. Following \cite{yan2025synthetic}, we use the depth network trained on the synthetic stage as a fixed teacher network, denoted as $\Phi_t$ in our framework. 
For clarity, the initial synthetic adaptation stage and the pose network training, which remain unchanged from~\cite{yan2025synthetic}, are omitted from Fig.~\ref{fig:framework}. 

\noindent \textbf{SAM Mask Generation} 
We leverage SAM to generate object-level information without any fine-tuning. For each input image $I_t$, we run SAM in its `segment-everything' mode to produce a set of $K$ binary masks, $M_{sam}^{raw} \in \{0, 1\}^{K \times H \times W}$. To obtain a partition of non-overlapping object-level segments, we apply a post-processing step: each pixel is assigned to the mask with the largest area among all candidate masks in $M_{sam}^{raw}$ that contain it.
This generates the final masks $M_{sam} \in \{0, 1\}^{N \times H \times W}$, where $N$ is the number of non-overlapping object-level regions and \( M_{sam,i} \in \{0,1\}^{H \times W} \) denotes the \( i \)-th region. All SAM masks are pre-generated before training, ensuring that SAM is only used offline. 
Additional details are provided in the supplementary material. 

\subsection{Object-Aware Representation Learning}
\label{subsectin:feature_level}
To inject object-level awareness into the depth network, we introduce segment-guided representation contrasting (SRC). 
Our SRC transfers segmentation priors from SAM through a regional prototype-based contrastive approach. 
It aims to encourage the depth network to infer geometric coherence, ensuring that pixels within the same segment are assigned structurally consistent depths. Specifically, SRC exploits SAM from a clustering perspective, contrasting the features of the depth network according to the given segments of the SAM. 

A key design choice of SRC is to perform the contrasting at the feature level rather than imposing direct constraints on the final depth. 
This is because SAM tends to over-segment a single object into multiple parts, and enforcing depth discontinuities at every segment boundary could lead to erroneous supervision. 
Instead, by transferring segmentation priors into the depth decoder's feature space, SRC allows the network to implicitly learn object-aware representations without explicitly altering the depths. 

\cref{fig:framework} illustrates the proposed SRC. 
The depth decoder produces a pyramid of multi-scale features. To incorporate segmentation cues into each decoder stage, we first resize the SAM masks using nearest interpolation to match the resolution of each feature level, denoted as $M_{sam}^{(s)}\in \{0,1\}^{N\times \frac{H}{2^s}\times \frac{W}{2^s}}$. 
At each scale $s$, we generate a prototype for each segment $M_{sam,i}^{(s)}$ by averaging the features of the pixels located on the segment:
\begin{equation}
\label{eq:ptototype}
\small
    \mathbf{pt}_i^{(s)} = \frac{1}{|M_{sam,i}^{(s)}|} \sum_{(x,y)\in M_{sam,i}^{(s)}} \mathbf{F}_{x,y}^{(s)},
\end{equation}
where $\mathbf{F}^{(s)}_{x,y}$ denotes the feature at position $(x,y)$ in scale $s$, and $\mathbf{pt}_i^{(s)}$ is the prototype of the $i$th segment. Note that we normalize the features/prototypes along the channel dimension before and after averaging, to ensure they are constrained to lie on the unit hypersphere. 

Then, we enforce each pixel feature to be close to the prototype of the segment it belongs to and distant from the prototypes of other segments. 
This contrastive objective encourages features within the same segment to form compact clusters, enabling the depth network to learn object-aware representations that lead to more consistent depth predictions. 
We formalize this contrasting process with the InfoNCELoss \cite{oord2018representation} as:
\begin{equation}
\label{eq:src}
\small
    L_{src} = -\sum_{s=1}^{S}\sum_{i=1}^{N}\sum_{(x,y)\in M_{sam,i}^{(s)}} \frac{\mathbf{F}_{x,y}^{(s)} \cdot \mathbf{pt}_{i}^{(s)} / T}{\sum_{j=1}^{N} \mathbf{F}_{x,y}^{(s)} \cdot \mathbf{pt}_j^{(s)} / T},
\end{equation}
where $S$ is the number of feature scales, $N$ is the total number of segments in the $M_{sam}$, and $T$ is the temperature. 

\subsection{Depth Learning with Object-Level Priors}
\label{subsection:depth level}
To further enhance the quality of predicted depth, we leverage object-level priors to improve depth learning signals. 
First, we propose to achieve object-aware smoothness by combining an adaptive regional outlier masking (AROM) strategy with a regional Gaussian likelihood loss. 
This combination suppresses intra-region outliers and maintains regional smoothness. 
Second, we develop an object-level reliability estimation (ORE) strategy for pseudo-label reweighting, which assesses depth reliability at the object level rather than the pixel level. 
These components together enforce region-consistent supervision and substantially improve the depth quality in diverse conditions. 

\noindent
\textbf{Object-aware Smoothness.}
\begin{figure}[!t]
    \centering
    \includegraphics[width=0.98\linewidth]{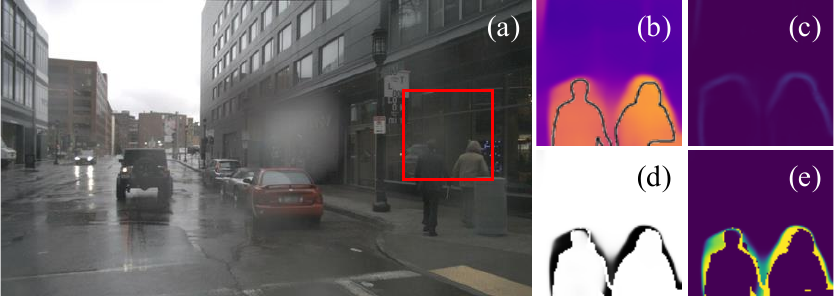}
    \caption{An example of why conventional boundary-aware smoothness loss performs poorly. (a) Original image. (b) Depth from $\Phi_t$ overlaid with the SAM boundary (black). (c) Weak and local gradients of the conventional boundary-aware smoothness loss. (d) Our outlier-suppressed mask. (e) Strong, non-local gradients from our proposed regional Gaussian likelihood loss guided by the outlier-suppressed mask.} 
    \vspace{-1em}
    \label{fig_boundary_smooth}
\end{figure}
This adaptation of the supervision signal comes from the motivation that object segmentation can be used to formulate a more precise smoothness constraint, thereby better regularizing the depth map. 
We begin by discussing the shortcomings of the conventional smoothness loss and then show how we mitigate those limitations with the help of SAM. 

\noindent \textit{Limitations of previous smoothness losses.}
Most existing methods define smoothness loss using first-order depth derivatives \cite{zhou2017unsupervised, godard2019digging, watson2021temporal, gasperini2023robust}. 
Given mean-normalized inverse depth $d^{*} = d/\overline{d}$, the conventional smoothness loss is:
\begin{equation}
\small
\label{eq:disparity_smoothness_loss}
L_{sm} \:=\: |\partial_x {d^{*}}|\:e^{-|\partial_x I|}  + \:|\partial_y d^{*}|\:e^{-|\partial_y I|},
\end{equation}
in which $e^{-|\partial I|}$ serves as the edge-aware weights to avoid penalties across object boundaries, where depth is not necessarily continuous. 
Such weights are usually derived from the image edges, which often coincide with object boundaries \cite{zhou2017unsupervised, godard2019digging}. However, the inherent limitations of the boundary-aware smoothness loss prevent it from leveraging more accurate object boundaries. 

One example is illustrated in \cref{fig_boundary_smooth}. 
As shown in \cref{fig_boundary_smooth}\hyperref[fig_boundary_smooth]{b}, the depth estimation fails around people, resulting in blurred depth and erroneously predicting part of the background building as a continuous foreground. 
In this case, the gradient of the boundary-aware smoothness loss mostly arises near the depth boundary (\cref{fig_boundary_smooth}\hyperref[fig_boundary_smooth]{c}), since the loss mainly focus on where the image is smooth ($e^{-|\partial I|}$ is large) and the depth exhibits disconunity ($|\partial d^{*}|$ is large). 
However, in regions where the depth is incorrectly estimated but locally smooth (e.g., mispredicted background between the object boundary and the erroneous depth boundary), the loss incorrectly assumes depth continuity ($|\partial d^{*}|$ is small) and fails to respond, resulting in weak gradients confined to a limited neighborhood around the depth boundary. 

\noindent \textit{Regional smoothness based on segment}. 
To address this limitation, we extend the smoothness formulation from object boundaries to object regions. 
Intuitively, the inaccurate depth values in \cref{fig_boundary_smooth}\hyperref[fig_boundary_smooth]{b} can be understood as regional outliers within the same object (background building). 
To achieve regional smoothness, we introduce an adaptive regional outlier masking (AROM) strategy together with a regional Gaussian likelihood loss. 
Specifically, AROM computes an outlier-suppressed mask $M_{out} \in [0,1]^{H\times W}$ that filters out depth outliers based on intra-object geometric consistency. 
To obtain $M_{out}$, for each object $M_{sam,i}$, we first calculate the mean $\overline{d_{t,i}}$ and standard deviation $\sigma_{t,i}$ of the inverse teacher depth $d_{t}$.  
Then we compute a dense deviation map $\delta\in \mathbb{R}^{H\times W}$, where each pixel's normalized deviation from its regional mean is obtained by:
\begin{equation}
\small
\delta = \sum_{i}^{N} M_{sam,i}\cdot\frac{|d_{t} - \overline{d_{t,i}}|}{\sigma_{t,i}}.
\end{equation} 

A fixed threshold on $\delta$ to identify outliers is suboptimal, as a small value tends to misclassify large but structurally consistent depth variations (e.g., ground or building surfaces) as outliers, while a large one fails to detect the outliers in homogeneous regions. 
We therefore introduce an adaptive threshold $\tau \in \mathbb{R}^{H\times W}$ that adjusts dynamically to each object's intrinsic depth variability. It relaxes the outlier criterion in high-variance regions (large $\sigma_{t,i}$) while enforcing a stricter one in homogeneous regions (small $\sigma_{t,i}$):
\begin{equation}
\small
    \tau
    = \sum_{i}^{N} M_{sam,i} \cdot
\left( \tau_0 + \lambda \sigma_{t,i} \right),
\end{equation}
where $\tau_0$ is a base threshold and $\lambda$ controls the sensitivity. 
The outlier-suppressed mask $M_{out}$ is generated by comparing deviation $\delta$ with the region-specific threshold $\tau$:
\begin{equation}
\small
    M_{out} = 1 - S_{\kappa}\!\big(S(\delta - \tau)\big),
\end{equation}
where $S(\cdot)$ is the sigmoid function and $S_{\kappa}(\cdot)$ is a temperature-controlled sharpening sigmoid that enhances the transition around 0.5. 
The resulting $M_{out}$ assigns low values to depth outliers  (\cref{fig_boundary_smooth}\hyperref[fig_boundary_smooth]{d}). 

With $M_{out}$, we introduce a regional Gaussian likelihood loss, which models the inverse teacher depth $d_t$ within each segment $M_{sam,i}$ as a Gaussian distribution characterized by $\overline{d_{t,i}}$ and $\sigma_{t,i}$. The loss encourages the inverse student depth $d_s$ intra-region smoothness by penalizing depth deviations from local statistics:
\begin{equation}
\small
\label{eq:gaussian_constraint}
L_{rgl}
=
(1-M_{out})\cdot
\sum_{i=1}^{N}
M_{sam,i}\cdot
\frac{(d_s - \overline{d_{t,i}})^2}{\sigma_{t,i}^{2}}.
\end{equation}
By weighting the loss with $1-M_{out}$, the supervision is concentrated on outliers. 
This formulation results in strong gradients on outlier regions (\cref{fig_boundary_smooth}\hyperref[fig_boundary_smooth]{e}) that enforce depth smoothness. 
Importantly, $M_{out}$ plays a dual role: it not only guides $L_{rgl}$, but also suppresses outlier-induced error propagation in the depth distillation loss (\cref{eq:distillation}). 

\noindent
\textbf{Object-Level Reliability.}
\begin{figure}[!t]
    \centering
    \includegraphics[width=0.99\linewidth]{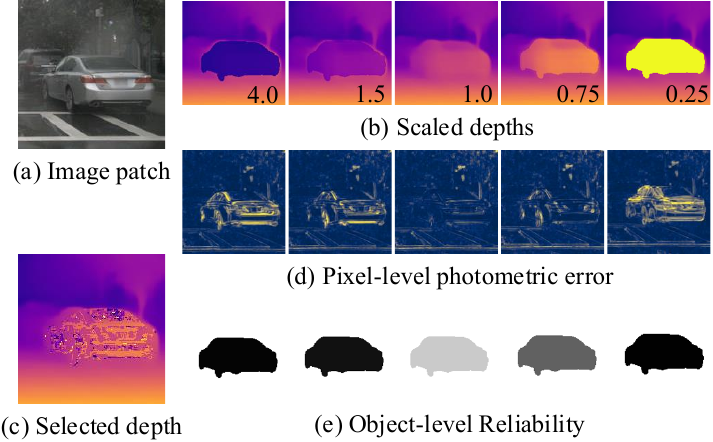}
    \caption{Pixel-level photometric error is insufficient for evaluating the object-level depth reliability. (a) Image patch. (b) Depth predictions from $\Phi_t$, with the car's depth scaled from 0.25 to 4.0. (c) The corresponding photometric error map, where brighter intensity denotes larger error, fails to capture the object-level depth reliability. (d) Depth map obtained by selecting, for each pixel, the scaled depth that minimizes the pixel-level photometric error. (e) Our object-level reliability correctly identifies the entire mispredicted car as an unreliable region.}
    \vspace{-1.2em}
    \label{fig_ins_obj}
\end{figure}
A widely adopted strategy for evaluating the reliability of pseudo-labels is to compute the pixel-wise photometric error \cite{saunders2023self, mao2024stealing}, assuming that low local errors correspond to trustworthy predictions. However, this local metric is misaligned with the nature of depth reliability, which typically exhibits strong object-level spatial correlations. 
As illustrated in \cref{fig_ins_obj}\hyperref[fig_ins_obj]{b} and~\hyperref[fig_ins_obj]{d}, scaling the depth of a car by different factors results in only localized deviations of the indicator (photometric error in \cref{eq:pe}), while the interior region remains falsely reliable. 
This is because pixels within the same object often share similar appearances, leading to chance alignments even when the object's depth is mispredicted. 
Consequently, pixel-level evaluation identifies only scattered unreliable pixels and fails to capture object-level unreliability. 

Furthermore, when selecting depths corresponding to the lowest indicator values as in \cite{saunders2023self, mao2024stealing}, many depths on incorrectly scaled factors are still considered trustworthy, producing incorrectly selected depths (see \cref{fig_ins_obj}\hyperref[fig_ins_obj]{c}). 
These findings reveal the inherent insufficiency of pixel-level indicators and motivate us to shift the focus from individual pixel evaluation to object-level reliability estimation. 
We define this as the object-level reliability estimation (ORE) strategy, which aims to assess the reliability of pseudo-labels within each object region for more effective supervision signals. 

Our approach begins by calculating the pixel-wise photometric error ($pe$) following \cite{godard2019digging, saunders2023self, mao2024stealing}:
\begin{equation}
\small
\label{eq:pe}
    pe = \alpha \cdot L_1(I_t,I_{t'}) + (1-\alpha) \cdot SSIM(I_t,I_{t'}).
\end{equation} 
where $I_t$ and $I_{t'}$ denote the target and reconstructed source images. 
We then exploit SAM masks to aggregate these pixel errors into a mean photometric error $\overline{pe_i}$ for each segment $M_{sam,i}$, following the same averaging rules as in \cref{eq:ptototype}. 
To obtain the final object-level reliability estimation, we compute the mean photometric error $\overline{pe}$ for the entire image and derive an object-level reliability map $R$ by comparing each object's error $\overline{pe_i}$ with $\overline{pe}$:
\begin{equation}
\small
R = \sum_i^{N}M_{sam,i} \cdot \text{exp}({- \beta \frac{max\{0, pe_i - \overline{pe}\}}{\overline{pe}}}),
\label{eq:ore}
\end{equation}
where $\beta$ is a scaling factor. 
To apply object-level reliability to the pixel-wise supervision, we construct a reliability-weighted map $W_{rel} = R+\epsilon$, where $\epsilon$ is a weight bias to ensure weak supervision, while preserving the relative emphasis on more reliable regions. 
Intuitively, as shown in \cref{fig_ins_obj}\hyperref[fig_ins_obj]{e}, objects whose errors deviate significantly from the global are assigned lower reliability. 

With $M_{out}$ and $W_{rel}$, we define our robust depth distillation loss as:
\begin{equation}
\small
\label{eq:distillation}
L_{d}
=
M_{out}\cdot W_{cst}\cdot W_{rel}\cdot
\big|\frac{D_s - D_t}{D_s}\big|,
\end{equation}
where $D_t$ and $D_s$ represent the depth predicted by $\Phi_t$ and $\Phi_s$, respectively. $W_{cst}$ denotes the consistency reweighting map from the baseline model \cite{yan2025synthetic}. 
The final training loss can be defined as:
\begin{equation}
    \small
    L_{total} = L_{d} + \lambda_1L_{src} + \lambda_2 L_{rgl} + L_{ext},
\end{equation}
where 
{\small $L_{ext}$} denotes the auxiliary losses adopted from~\cite{yan2025synthetic}, and more details are shown in supplementary. 

\begin{table*}[!t] 
\renewcommand{\arraystretch}{1.0}

\centering
\footnotesize
\setlength{\tabcolsep}{3.5pt}  
\begin{tabular}{l|c|cccc|cccc|cccc} 
\toprule 
\multirow{2}{*}{Method} & \multicolumn{1}{c|}{Test} & \multicolumn{4}{c|}{day-clear -- nuScenes} & \multicolumn{4}{c|}{night -- nuScenes} & \multicolumn{4}{c}{day-rain -- nuScenes}\\
     &frames& \cellcolor{orange!30}AbsRel$\downarrow$ &\cellcolor{orange!30} SqRel$\downarrow$ & \cellcolor{orange!30} RMSE$\downarrow$ & $ \cellcolor{up} \delta_1\uparrow $ & \cellcolor{orange!30} AbsRel$\downarrow$ & \cellcolor{orange!30} SqRel$\downarrow$ &  \cellcolor{orange!30} RMSE$\downarrow$ &\cellcolor{up} $ \delta_1\uparrow $ & \cellcolor{orange!30} AbsRel$\downarrow$ & \cellcolor{orange!30} SqRel$\downarrow$ &  \cellcolor{orange!30} RMSE$\downarrow$ &\cellcolor{up} $ \delta_1\uparrow $    \\  
\midrule
Monodepth2 \cite{godard2019digging} & 1 &0.1333 &1.820 &6.459 &85.88 &0.2419 &2.776 &10.922 &58.17 & 0.1572 &2.273 &7.453 & 79.49 \\
R4Dyn$^{\textbf{\dag}}$ (Radar) \cite{gasperini2021r4dyn} & 1 &0.1259 &1.661 &6.434 &\textbf{86.97} &0.2194 & 2.889  &10.542 &62.28 &0.1337 &1.938 &7.131 &83.91 \\
RNW \cite{wang2021regularizing} & 1 &0.2872 &3.433  &9.185 &56.21 &0.3333 &4.066 &10.098 &43.72 &0.2952 &3.796 &9.341 &57.21 \\
 robust-depth \cite{saunders2023self}  & 1 &0.1436 &1.862 &6.802 &82.95 &0.2101 &2.691 &8.673 &69.58 &0.1458 &1.891 &7.371 &80.21 \\
  md4all-AD \cite{gasperini2023robust}   & 1 &0.1523 &2.141 &6.853 &83.11 &0.2187 &2.991 &9.003 &68.84 &0.1601 &2.259 &7.832 &78.97 \\
  md4all-DD \cite{gasperini2023robust} & 1 &0.1366 &1.752 &6.452 &84.61 &0.1921 & 2.386 &8.507 &71.07 &0.1414 &1.829 &7.228 &80.98 \\
  DM-MDE \cite{tosi2024diffusion} & 1 &0.1280 &- &6.449 &84.03 &0.1910 &- &8.433 &\underline{71.14} &0.1390 &- &7.129 &81.36 \\
  Syn2Real-Depth \cite{yan2025synthetic} & 1 &\underline{0.1239} & \underline{1.454} &\underline{6.214} &85.66 &\underline{0.1792} &\underline{1.896}&\textbf{7.949} &71.08 &\underline{0.1331} &\underline{1.616} &\underline{6.926} &\underline{81.68} \\
 \textbf{Ours-Single} & 1 & \textbf{0.1172} &\textbf{1.357} &\textbf{6.121} &\underline{86.13} &\textbf{0.1742} &\textbf{1.865} &\underline{7.966} &\textbf{72.52} &\textbf{0.1298} &\textbf{1.612} &\textbf{6.801} &\textbf{82.30} \\
\midrule
 Manydepth \cite{watson2021temporal} & 2(-1,0) &0.1213 &1.699 &6.183 &\textbf{87.05} &0.2442 &3.347 &11.38 &57.04 &0.1347 &1.730 &7.110 &81.24 \\
 Syn2Real-Depth \cite{yan2025synthetic} & 2(-1,0) &\underline{0.1170} &\underline{1.371} &\underline{6.023} &86.66 & \underline{0.1713} & \underline{1.779} &\textbf{7.689} &\underline{73.16} &\underline{0.1266} &\underline{1.532} &\underline{6.654} &\underline{82.86} \\
\textbf{Ours-Multi} & 2(-1,0) &\textbf{0.1131} &\textbf{1.281} &\textbf{5.992} &\underline{86.95} & \textbf{0.1667} & \textbf{1.775} &\underline{7.729} &\textbf{73.95} &\textbf{0.1236} &\textbf{1.521} &\textbf{6.613} &\textbf{83.53} \\
\bottomrule 
\end{tabular}
\caption{Quantitative comparison on nuScenes \cite{caesar2020nuscenes} under different conditions (day-clear, nighttime, and day-rain). $\textbf{\dag}$ denotes the method \cite{gasperini2021r4dyn} that leverages additional Radar information. `Test frames' specifies whether the evaluation is conducted in a multi-frame or single-frame setting. The
best and the second best results are highlighted in \textbf{bold} and \underline{underline}, respectively.}
\label{tab:nuscenes}
\vspace{-0.3cm}
\end{table*}

\begin{table*}[!tb] 
\renewcommand{\arraystretch}{1.0}
\centering
\footnotesize
\resizebox{.82\linewidth}{!}{
\begin{tabular}{l|c|cccc|cccc} 
\toprule 
\multirow{2}{*}{Method } & \multicolumn{1}{c|}{Test} & \multicolumn{4}{c|}{day -- Robotcar} & \multicolumn{4}{c}{night -- Robotcar}\\

                        & frames & \cellcolor{orange!30} AbsRel$\downarrow$ & \cellcolor{orange!30} SqRel$\downarrow$ & \cellcolor{orange!30}  RMSE$\downarrow$ & $\cellcolor{up} \delta1\uparrow $ &\cellcolor{orange!30}  AbsRel$\downarrow$ & \cellcolor{orange!30} SqRel$\downarrow$ &  \cellcolor{orange!30} RMSE$\downarrow$ & $ \cellcolor{up}\delta1\uparrow $  \\  
\midrule
   Monodepth2 \cite{godard2019digging}                  &1 &0.1209 &0.723 &3.335 &86.61 &0.3909 &3.547 &8.227 &22.51 \\
  DeFeatNet \cite{spencer2020defeat}     &1 &0.2470 &2.980 &7.884 &65.00 &0.3340 &4.589  &8.606 &58.60 \\
  ADIDS \cite{liu2021self}     &1&0.2390 &2.089 &6.743 &61.40 &0.2870 &2.569  &7.985 &49.00 \\
   RNW \cite{wang2021regularizing}               &1 &0.2970 &2.608  &7.996 &43.10 &0.1850 &1.710 &6.549 &73.30  \\
   WSGD  \cite{wang2021regularizing}                   &1&0.1760 &1.603 &6.036 &75.00 &0.1740 &1.637 &6.302 &75.40 \\
   robust-depth \cite{saunders2023self}            & 1 &0.1225 &0.776 &3.302 &85.84 &0.1333 &0.895 &3.756 &85.11  \\
   md4all-DD \cite{gasperini2023robust}       & 1 &0.1128 &0.648 & 3.206 &87.13 &0.1219 &0.784 &3.604 &84.86 \\
   DM-MDE \cite{tosi2024diffusion}       & 1 &0.1190 &0.728 &3.287 &87.17 &0.1290 &0.751 &3.661 &83.86 \\
   \midrule
   Manydepth \cite{watson2021temporal}     & 1 &0.1200 &0.802 &3.413 &87.46 &0.4974 &5.517 &10.21 &14.87 \\
      Syn2Real-Depth\cite{yan2025synthetic}       & 1 &\underline{0.1063} &\underline{0.646} &\underline{3.171} &\underline{88.29} &\underline{0.1103} &\underline{0.693} &\underline{3.455} &\underline{86.15}  \\
   \textbf{Ours-Single}                  & 1 &\textbf{0.1006} &\textbf{0.595} &\textbf{3.022} &\textbf{88.90} &\textbf{0.1066} &\textbf{0.643} &\textbf{3.291} &\textbf{87.03}  \\
\bottomrule 
\end{tabular}}
\caption{Quantitative comparison on the Robotcar \cite{maddern20171} test set under daytime and nighttime conditions.}
\label{tab:robotcar}
\vspace{-0.4cm}
\end{table*}
\vspace{-0.8em}
\section{Experiments}
\begin{figure*}[t]
    \centering
    \includegraphics[width=0.96\linewidth]{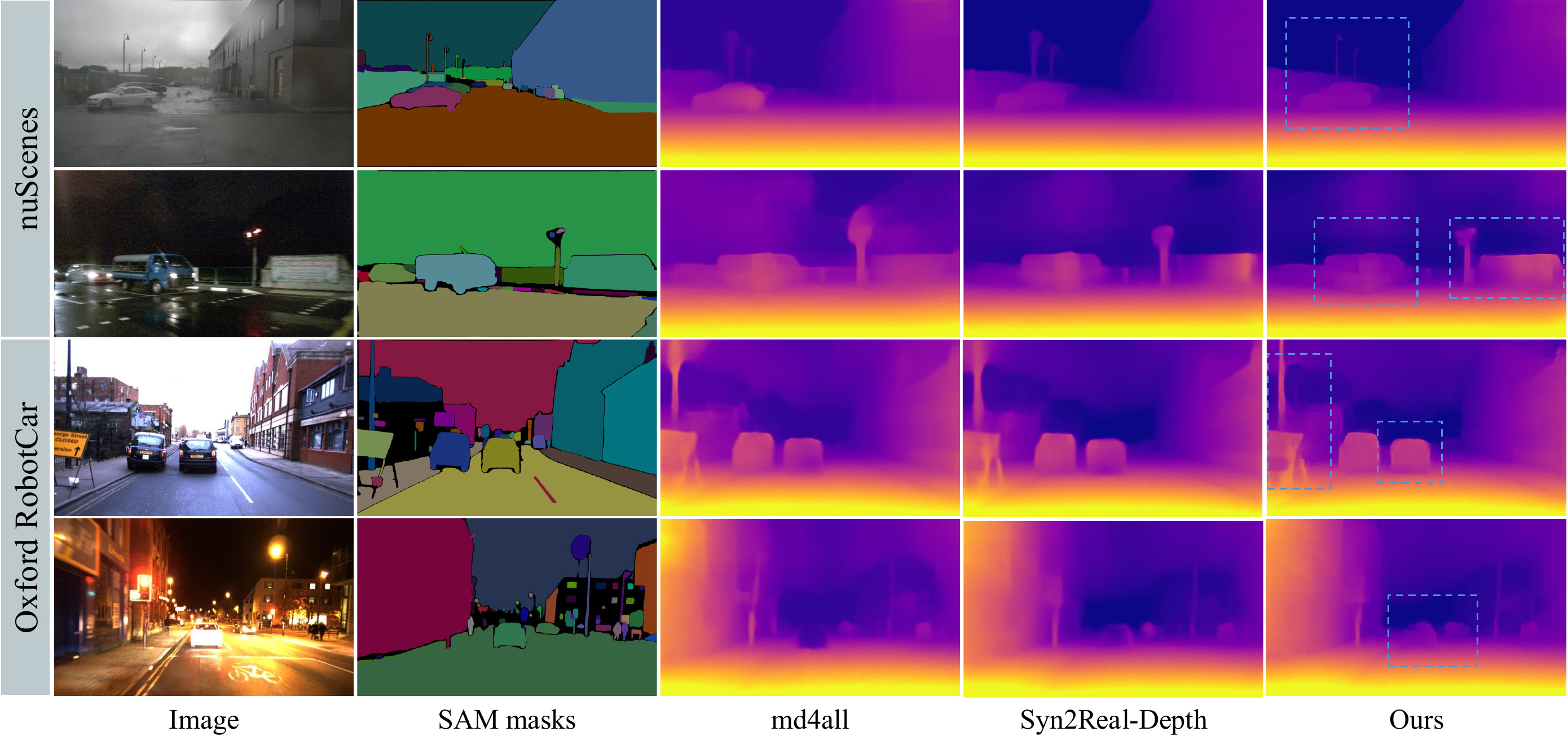}
    \caption{Qualitative comparison of depth predictions under diverse conditions. The first two rows show results on nuScenes \cite{caesar2020nuscenes} (rain and nighttime), and the last two rows show results on Robotcar \cite{maddern20171} (daytime and nighttime).}
    \label{fig:qualitative}
\end{figure*}
\vspace{-0.3em}
\subsection{Implement Details}
\vspace{-0.3em}
Both our teacher and student depth networks adopt the same architecture \cite{watson2021temporal} as in~\cite{yan2025synthetic}. 
In our implementation, we use the pretrained teacher model provided by~\cite{yan2025synthetic}, which is trained on synthetic data, without any fine-tuning. 
The student network $\Phi_s$ is initialized from the teacher $\Phi_t$ and trained for 10 epochs following~\cite{yan2025synthetic}. 
The model is implemented in Pytorch \cite{paszke2019pytorch} and optimized using Adam~\cite{kingma2014adam} with a batch size of 10. 
The learning rate is initialized to $8 \times e^{-5}$ and decayed by a factor of 0.5 every 5 epochs. 
For the Segment Anything Model \cite{kirillov2023segment}, we use the off-the-shelf default ViT-H pretrained model. Additional implementation details are provided in the supplementary material. 

\vspace{-0.3em}
\subsection{Dataset and Evaluation Metric}
\vspace{-0.2em}
We evaluate our method on the nuScenes \cite{caesar2020nuscenes} and Oxford Robotcar \cite{maddern20171} datasets and follow the same training data schedules as \cite{yan2025synthetic}. 
NuScenes \cite{caesar2020nuscenes} is a comprehensive benchmark with real driving data under daytime, nighttime, and rainy conditions. 
In our experiments, we utilize 21,476 images for training, including 15,129 daytime, 3,796 rainy, and 2,551 nighttime frames. 
The evaluation set consists of 6,019 consecutive frame pairs: 4,449 daytime, 1,088 rainy, and 602 nighttime. 
We report results in both single-frame and multi-frame settings, and all metrics are computed for ground truth within 0.1--80\,m. 
For the Oxford RobotCar \cite{maddern20171} dataset, following WSGD \cite{vankadari2023sun}, we use 17,790 daytime and 19,162 nighttime images for training, and 1,411 single images for testing (702 daytime images and 709 nighttime images). 
Evaluation is performed under the single-frame setting, with metrics computed within 0.1--50\,m. 

We employ commonly adopted metrics \cite{godard2019digging, wang2021regularizing, gasperini2023robust, tosi2024diffusion,yan2025synthetic} for depth estimation, including absolute relative error (AbsRel), root mean square error (RMSE), squared relative error (SqRel), and threshold accuracy ($\delta_1$). 

\subsection{Quantitative Results}
\label{sec:quantitative_exp}
Our quantitative testing results across different datasets are shown in \cref {tab:nuscenes,tab:robotcar}. 
Our method significantly outperforms state-of-the-art unsupervised methods on both nuScenes \cite{caesar2020nuscenes} and Oxford Robotcar \cite{maddern20171} datasets on almost all evaluation metrics. 
On the nuScenes dataset \cite{caesar2020nuscenes}, which covers diverse conditions including day-clear, night, and day-rain. Following \cite{yan2025synthetic}, we evaluate both single-frame (identical inputs) and multi-frame (temporally adjacent inputs) settings. Our method consistently demonstrates superior performance across almost all metrics and conditions. 
Compared with recent methods trained on synthetic data \cite{gasperini2023robust, tosi2024diffusion}, our method significantly outperforms them. When compared to the real-data-trained method \cite{yan2025synthetic}, our model achieves consistent improvements across almost all metrics under both single-frame and multi-frame settings, with average relative gains of $2.8\%$, $2.7\%$, $0.6\%$, and $0.9\%$ on AbsRel, SqRel, RMSE, and $\delta_1$, respectively, averaged over all conditions and both single- and multi-frame settings. 

In \cref{tab:robotcar}, we present a comparative analysis on the Robotcar dataset \cite{maddern20171}, evaluating the performance of previous methods against our method. Since the Oxford Robotcar test set contains only single-frame samples, we evaluate under the single-frame setting. 
Our method achieves substantial improvements of $4.4\%, 7.6\%, 4.7\%, 0.9\%$ on AbsRel, SqRel, RMSE, and $\delta_1$, respectively, compared with ~\cite{yan2025synthetic}. 
These results demonstrate the effectiveness of our method, which leverages SAM priors to improve the depth quality and robustness in self-supervised depth estimation under diverse conditions. 

\subsection{Qualitative Experimental Results}
\vspace{-0.5em}
\label{sec:qualitative_exp}
Our qualitative results across various adverse conditions are shown in \cref{fig:qualitative}. 
As the Oxford Robotcar dataset already includes clear day scenes, we omit the day-clear cases from nuScenes and show representative samples from both datasets. More qualitative examples are available in the supplementary material. 
Since DM-MDE \cite{tosi2024diffusion} is unavailable, we compare with the latest methods \cite{gasperini2023robust, yan2025synthetic}. 
Under the day-rain condition (nuScenes, Row1), images often exhibit low contrast and texture degradation, leading md4all~\cite{gasperini2023robust} and Syn2Real-Depth~\cite{yan2025synthetic} to blur object boundaries. In contrast, our method maintains sharper structures and accurately recovers the spatial relationships among objects. 
Under the clear daytime condition (Oxford Robotcar, Row 3), our method predicts high-quality depth consistent with scene geometry, producing accurate depth for the car and finer delineation of poles and street signs. 
Nighttime scenes pose additional challenges, such as limited illumination (nuScenes, Row 2) and strong glare (nuScenes, Row 4). Competing methods often fail to recover consistent and sharp depth in these cases. 
However, our method effectively preserves depth consistency and boundary sharpness, suppresses reflection-induced artifacts, and clearly defines the foreground while accurately capturing the depth of the more distant vehicles. 
These visual comparisons highlight our framework’s robustness across diverse illumination and weather conditions, particularly in preserving object boundaries and maintaining depth consistency under different challenging scenes. 

\begin{table*}[!thb] 
\renewcommand{\arraystretch}{1.0}
\centering
\footnotesize
\resizebox{.92\linewidth}{!}{
\begin{tabular}{cccc|cccc|cccc} 
\toprule 
\multirow{2}{*}{SRC} & \multirow{2}{*}{AROM}&\multirow{2}{*}{$L_{rgl}$}&\multirow{2}{*}{ORE}&\multicolumn{4}{c|}{night -- nuScenes} & \multicolumn{4}{c}{day-rain -- nuScenes}\\

                          & & &  & \cellcolor{orange!30}AbsRel$\downarrow$ & \cellcolor{orange!30}SqRel$\downarrow$ & \cellcolor{orange!30} RMSE$\downarrow$ &\cellcolor{up} $ \delta1\uparrow $ &\cellcolor{orange!30} AbsRel$\downarrow$ &\cellcolor{orange!30} SqRel$\downarrow$ &\cellcolor{orange!30}  RMSE$\downarrow$ &\cellcolor{up} $ \delta1\uparrow $  \\  
\midrule
    \checkmark &&&&0.1725 &1.833 &7.828 &73.21&0.1270 &1.623 &6.710 &82.82         \\
    &\checkmark&&& 0.1711& 1.806& 7.774& 73.24& 0.1263 & 1.612 & 6.681 & 83.03 \\
    &&\checkmark&& 0.1734& 1.841& 7.819& 73.14& 0.1289 & 1.657&6.794 & 82.87 \\
           &\checkmark &\checkmark&&  0.1699 & 1.804 & 7.756 & 73.67 & 0.1259 & 1.603 & 6.639 & 83.17     \\
           &\checkmark&\checkmark&\checkmark&0.1673 & 1.798 & 7.736 & 73.89 & 0.1249 & 1.608 & 6.621 & 83.45     \\
        \checkmark&\checkmark&\checkmark&\checkmark& \textbf{0.1667} &\textbf{1.775} & \textbf{7.729} & \textbf{73.95} &\textbf{0.1236}  &\textbf{1.521} &\textbf{6.613}& \textbf{83.53}   \\
  
\bottomrule 
\end{tabular}}
\caption{Ablations study of our components on nuScenes \cite{caesar2020nuscenes}. The checkmark indicates the inclusion of the corresponding component.}
\label{table:ablation}
\vspace{-0.3cm}
\end{table*}

\begin{table}[t]
    \centering
    \resizebox{0.95\linewidth}{!}{
    \begin{tabular}{c|cc|cc}
    \toprule
         \multirow{2}{*}{Smoothness Loss}& \multicolumn{2}{c|}{night -- nuScenes}  &\multicolumn{2}{c}{day-rain -- nuScenes} \\
          &\cellcolor{orange!30}AbsRel$\downarrow$ &\cellcolor{up} $ \delta1\uparrow $ &\cellcolor{orange!30}AbsRel$\downarrow$ &\cellcolor{up}$ \delta1\uparrow $ \\
          \midrule
         Image-edge-aware&0.1735 &73.15&0.1280&82.75 \\
         SAM-boundary-aware& 0.1732&73.17&0.1273& 82.78\\
         \midrule
         AROM \&$L_{rgl}$ (Ours)&\textbf{0.1699} &\textbf{73.67}&\textbf{0.1259}&\textbf{83.17} \\
    \bottomrule
    \end{tabular}}
    \caption{Ablation study on different smoothness loss definitions, SAM-boundary-aware smoothness loss indicates replacing $I$ with $M_{sam}$ in \cref{eq:disparity_smoothness_loss}.}
    \label{tab:smoothness}
    \vspace{-0.5em}
\end{table}

\begin{table}[t]
\renewcommand{\arraystretch}{0.9}
    \centering
    \resizebox{0.85\linewidth}{!}{
    \begin{tabular}{c|cc|cc}
    \toprule
         \multirow{2}{*}{Method}& \multicolumn{2}{c|}{night -- nuScenes}  &\multicolumn{2}{c}{day-rain -- nuScenes} \\
          &\cellcolor{orange!30}AbsRel$\downarrow$ &\cellcolor{up} $ \delta1\uparrow $ &\cellcolor{orange!30}AbsRel$\downarrow$ &\cellcolor{up}$ \delta1\uparrow $ \\
          \midrule
         SRC on $D_s$& 0.1694&73.68&0.1261&83.05 \\
         SRC (Ours)&\textbf{0.1667} &\textbf{73.95}&\textbf{0.1236}&\textbf{83.53}\\
    \bottomrule
    \end{tabular}}
    \caption{Ablation study on different implementations of \cref{eq:src}.}
    \label{tab:sasc}
    \vspace{-1.5em}
\end{table}

\vspace{-0.3em}
\section{Ablations}
\vspace{-0.3em}
\label{sec:ablations}
We do extensive ablation studies to analyze the effectiveness of our major contributions. 

\noindent
\textbf{SRC.} We ablate the application strategy for SRC in \cref{tab:sasc}, finding that applying it directly to predicted depth degrades performance. As discussed in \cref{subsectin:feature_level}, this is because some SAM mask fragmentation introduces noise by forcing discontinuities on the same object. We also analyze it in \cref{table:ablation}. SRC plays a critical role in our full model when combined with the other components. It effectively enhances the feature when the predicted depth tends to align with SAM masks, which is not the case when it is solely applied. 

\noindent
\textbf{AROM \& $\mathbf{L_{rgl}}$.} As shown in \cref{tab:smoothness}, our proposed AROM combined with $L_{rgl}$ significantly outperforms the conventional edge-aware smoothness loss based on image edges or SAM boundaries, which is consistent with our analysis in \cref{subsection:depth level}. 
We also ablate their individual roles in \cref{table:ablation}. Removing AROM prevents the model from suppressing pseudo-label outliers in the depth distillation loss (\cref{eq:distillation}), while removing $L_{rgl}$ makes the model lack regional supervision needed to enforce smoothness. 
Consequently, either component alone leads to inferior performance. 

\noindent
\textbf{ORE.} As shown in \cref{table:ablation}, by incorporating object-level reliability through ORE, the depth distillation loss becomes more effective, enabling the model to progressively improve its performance under real-world diverse conditions.

\vspace{-0.6em}
\section{Conclusion}
\vspace{-0.4em}
\label{sec:conclusion}
This work presents RoSAMDepth, a novel robust self-supervised monocular depth estimation framework that leverages object-level priors from SAM during training. Our method integrates three key components: segment-guided representation contrasting (SRC) for structural awareness, adaptive regional outlier masking (AROM) with regional Gaussian likelihood loss for consistent depth, and object-level reliability estimation (ORE) for reliability-aware learning. Experiments on challenging datasets demonstrate consistent improvements over existing methods across diverse conditions.

{
    \small
    \bibliographystyle{ieeenat_fullname}
    \bibliography{main}
}

\clearpage
\appendix
\setcounter{figure}{0}
\setcounter{table}{0}
\setcounter{equation}{0}
\renewcommand{\thefigure}{S\arabic{figure}}
\renewcommand{\thetable}{S\arabic{table}}
\renewcommand{\theequation}{S\arabic{equation}}
\renewcommand{\theHfigure}{supp.\arabic{figure}}
\renewcommand{\theHtable}{supp.\arabic{table}}
\renewcommand{\theHequation}{supp.\arabic{equation}}
\maketitlesupplementary

\section{Methods Details}
\subsection{SAM Mask Generation}
\begin{figure*}[t]
            \centering
    \includegraphics[width=0.75\linewidth]{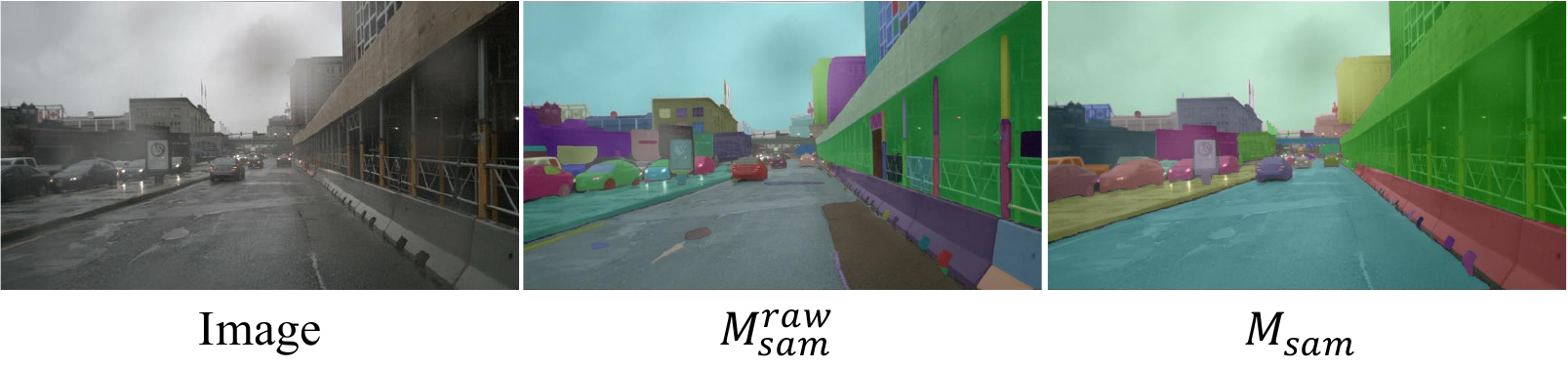}
    \caption{Example of the SAM masks generation.}
    \label{fig: mask_process}
    \vspace{-1.5em}
\end{figure*}

SAM often produces a hierarchical set of masks for a single object, detecting not only the entire object but also its constituent parts (e.g., wheels, doors, windows, lights, and even bumpers for a car, as shown in \cref{fig: mask_process}). 
While detailed, this hierarchy introduces redundancy and potential noise for tasks requiring object-level information. 
To address this, we adopt a simple yet effective strategy. 
We assume that in overlapping masks, the mask with the largest spatial extent represents the object instance, while smaller nested masks represent its parts. 

Formally, let $M_{sam}^{raw}\in \{0,1\}^{K\times H \times W} $ denote the set of $K$ binary masks produced by SAM for an input image $I$, where  $M^{raw}_{sam,i} \in \{0, 1\}^{H \times W}$ is the $i$-th mask. 
Specifically, we iterate through masks with an area below a threshold (set to $10^5$ pixels, approx. $200 \times 500$). 
If a small mask is significantly covered by a larger mask (overlap ratio $> 0.75$), it is considered a sub-part and is merged into the larger parent mask. As shown in \cref{fig: mask_process}, the resulting unified segmentations serve as robust object-level priors. 
These object-level priors ensure that the network achieves consistent and accurate depth predictions across the entire object instance. 

\subsection{Training Framework}
We build upon the Syn2Real-Depth framework~\cite{yan2025synthetic} to investigate how object-level priors from SAM~\cite{kirillov2023segment} can enhance robust self-supervised depth estimation in diverse real-world scenes. 
Syn2Real-Depth serves as an ideal baseline due to its real adaptation stage designed for diverse real data. 
The framework employs ManyDepth~\cite{watson2021temporal} as the depth backbone, originally defining three domain-specific models: $\Phi_{day}$ (trained on real daytime data), $\Phi_{syn}$ (trained on diverse synthetic data), and $\Phi_{real}$ (trained on diverse real data). 
The pose network follows the standard architecture described in~\cite{godard2019digging, watson2021temporal}. 
Since our contribution focuses on integrating SAM-based guidance into the real-world adaptation stage, we simplify the notation to better reflect our training paradigm. 
Specifically, we designate $\Phi_{syn}$ as the initial teacher $\Phi_t$, and $\Phi_{real}$ as the student $\Phi_s$, which is fine-tuned on real data with object-level priors. 
\subsection{Implementation Details}
Our model is trained on a single NVIDIA 4090D GPU, taking about 10 hours. 
Regarding the hyperparameters, we set the temperature $T = 1.0$ and the number of feature scales $S = 3$ for $L_{src}$. 
For the temperature-controlled sharpening sigmoid function $S_{\kappa}(\cdot)$, the sharpening temperature is set to $\kappa = 15.0$. 
In the object-level reliability estimation (ORE) strategy, we employ a scaling factor $\beta = 1.0$ and a bias term $\epsilon = 1.0$. 
Finally, the balancing weights for the total objective $L_{total}$ are set to $\lambda_1 = 0.05$ and $\lambda_2 = 0.2$. 

\subsection{Supervision Details}
To ensure robust performance under adverse conditions, our training on real data inherits the auxiliary losses from Syn2Real-Depth~\cite{yan2025synthetic}. Specifically, we incorporate the external loss term $L_{ext}$, which comprises the K-L divergence ($L_{dis}^{KL}$), the cost volume learning loss ($L_{cv}$), and the supervision for pose ($L_T$), following the exact formulation in~\cite{yan2025synthetic}. We retain these components unchanged to leverage their proven efficacy in the real data. 
This allows us to isolate the source of performance gains, demonstrating that our proposed object-level integration provides additive improvements over a strong baseline. 

\section{Additional Results}
\subsection{Ablation on Number of Feature Scales \texorpdfstring{$S$}{S}}
We analyze the sensitivity of our method to the number of feature scales used in \cref{eq:src} and the results are presented in \cref{tab:suup_src}. We ultimately set $S = 3$ to balance efficiency and accuracy, as including the highest-resolution feature map introduces significant computational overhead without yielding noticeable performance gains. 
\begin{table}[t]
\renewcommand{\arraystretch}{0.9}
    \centering
    \resizebox{0.85\linewidth}{!}{
    \begin{tabular}{c|cc|cc}
    \toprule
         \multirow{2}{*}{Method}& \multicolumn{2}{c|}{night -- nuScenes}  &\multicolumn{2}{c}{day-rain -- nuScenes} \\
          &\cellcolor{orange!30}AbsRel$\downarrow$ &\cellcolor{up} $ \delta1\uparrow $ &\cellcolor{orange!30}AbsRel$\downarrow$ &\cellcolor{up}$ \delta1\uparrow $ \\
          \midrule
          S=1& 0.1674&73.88&0.1250&83.45 \\
         S=2& 0.1671&73.91&0.1244&83.47 \\
          S=3& 0.1667&\textbf{73.95}&0.1236&\textbf{83.53} \\
         S=4&\textbf{0.1665} &\textbf{73.95}&\textbf{0.1234}&\textbf{83.55}\\
    \bottomrule
    \end{tabular}}
    \caption{Ablation study on the number of feature scales ($S$) used in \cref{eq:src}.}
    \vspace{-1.5em}
    \label{tab:suup_src}
\end{table}

\subsection{More Qualitative Examples}
\begin{figure*}[t]
        \centering
    \includegraphics[width=1\linewidth]{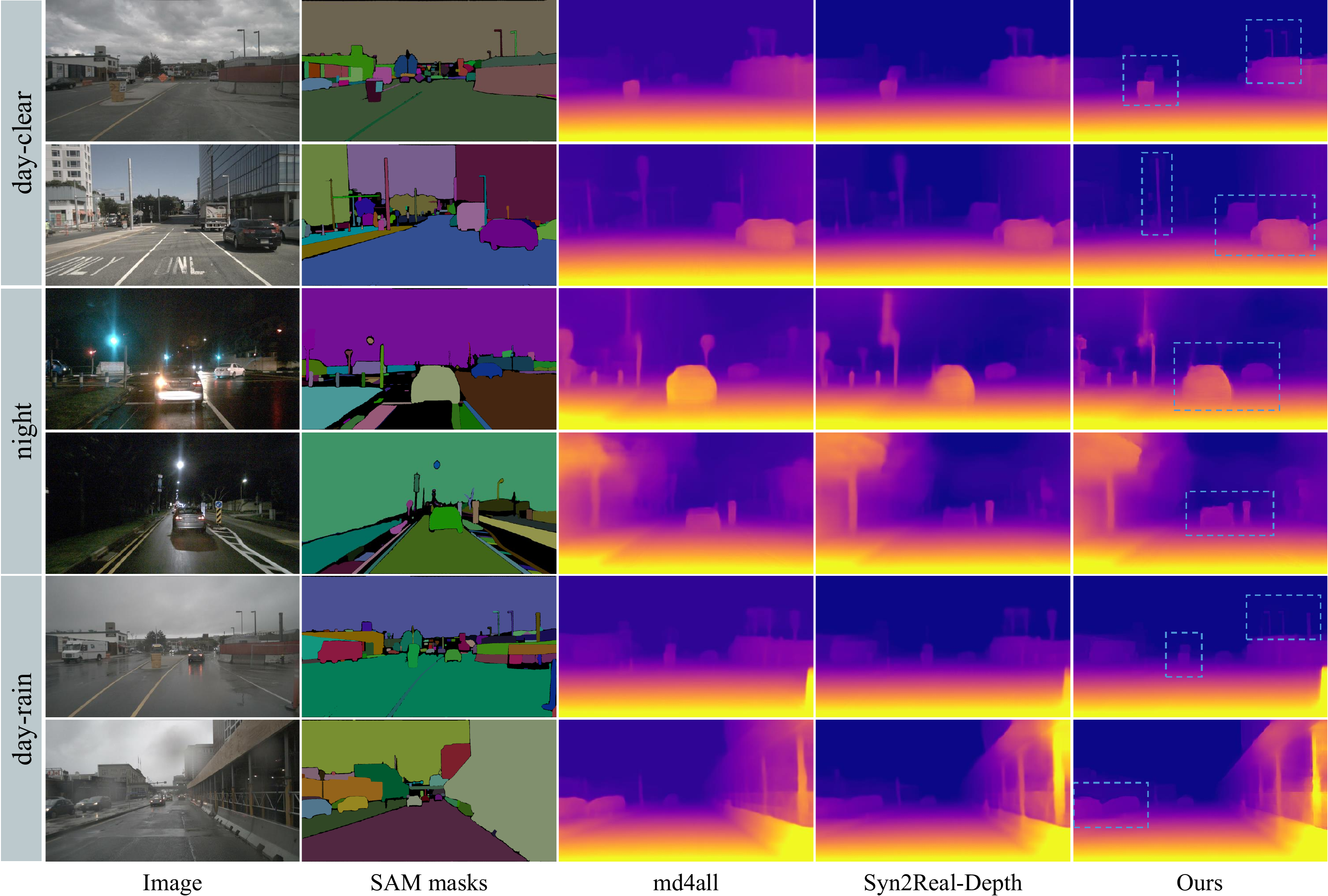}
    \caption{More qualitative results on nuScenes \cite{caesar2020nuscenes}}
    \label{fig:suup_nuscenes}
\end{figure*}

\begin{figure*}
        \centering
    \includegraphics[width=1\linewidth]{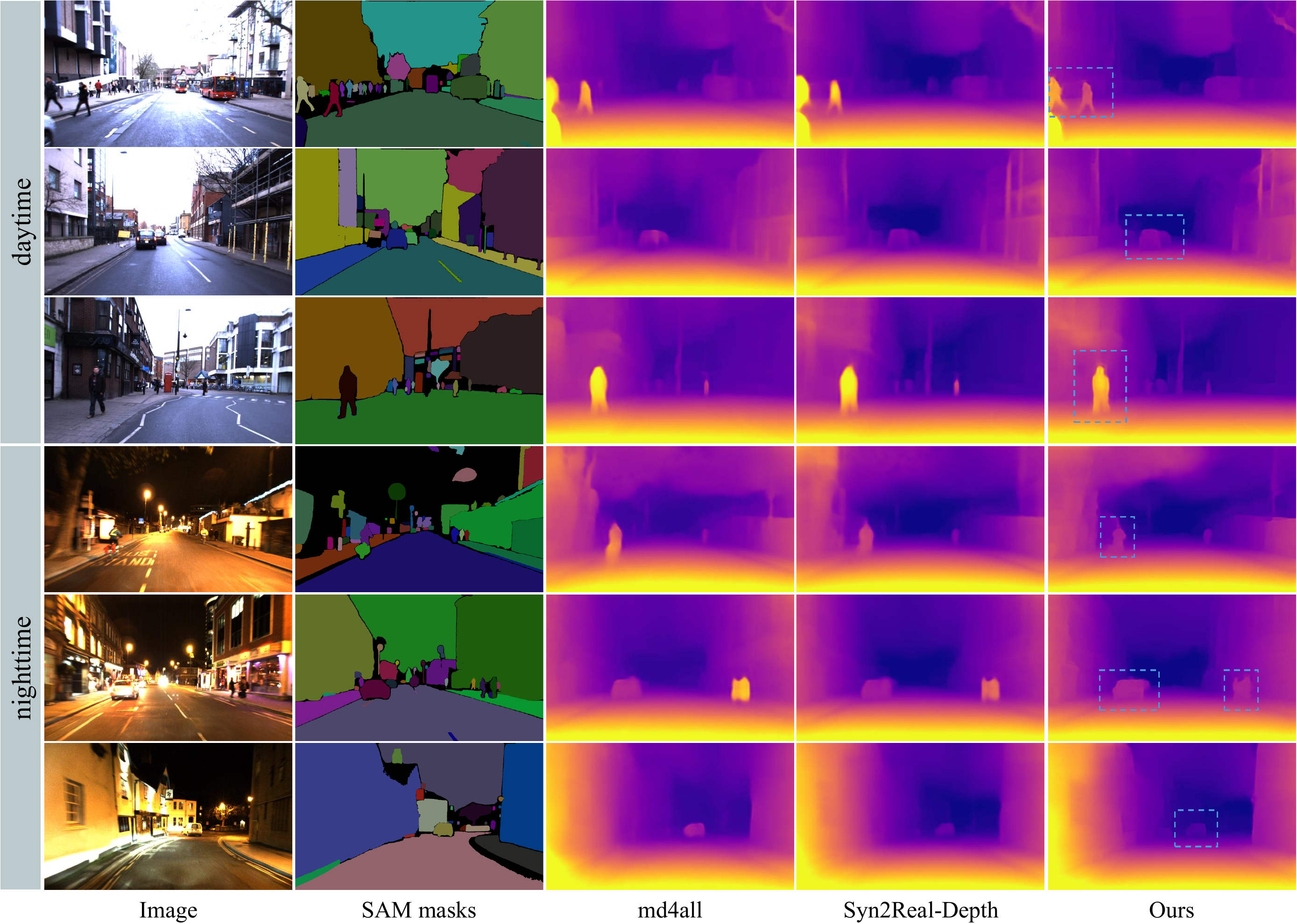}
    \caption{More qualitative results on Robotcar \cite{maddern20171}}
    \label{fig:suup_robotcar}
\end{figure*}

As supplementary to the main paper, we provide more qualitative results on nuScenes \cite{caesar2020nuscenes} (day-clear, night, day-rain), Robotcar \cite{maddern20171} (daytime, nighttime), as shown in \cref{fig:suup_nuscenes,fig:suup_robotcar}.

\end{document}